\documentclass[sigconf]{acmart}
\AtBeginDocument{%
  }

\setcopyright{acmcopyright}
\copyrightyear{2018}
\acmYear{2018}
\acmDOI{XXXXXXX.XXXXXXX}

\acmConference[MM '23]{The 31st ACM International Conference on Multimedia}{October 28 -- November 3, 2023}{Ottawa, Canada}
\acmPrice{15.00}
\acmISBN{978-1-4503-XXXX-X/18/06}

\usepackage{multirow}
\usepackage{multicol}
\usepackage{caption}
\usepackage{subcaption}
\usepackage{cleveref}
\crefname{section}{Sec.}{Secs.}
\Crefname{table}{Table}{Tables}
\crefname{table}{Table}{Tables}
\crefname{figure}{Figure}{Figures}

\usepackage{tikz}
\newcommand{\tikzxmark}{%
\tikz[scale=0.23] {
    \draw[line width=0.7,line cap=round] (0,0) to [bend left=6] (1,1);
    \draw[line width=0.7,line cap=round] (0.2,0.95) to [bend right=3] (0.8,0.05);
}}
\newcommand{\tikzcmark}{%
\tikz[scale=0.23] {
    \draw[line width=0.7,line cap=round] (0.25,0) to [bend left=10] (1,1);
    \draw[line width=0.8,line cap=round] (0,0.35) to [bend right=1] (0.23,0);
}}
\acmSubmissionID{3527}
\begin{document}

\title{Multiview Transformer: Rethinking Spatial Information in Hyperspectral Image Classification}

\author{Jie Zhang}
\email{Jiezh1997@gmail.com}
\affiliation{
  \institution{University of Macau}
  \country{Macau, China}
}

\author{Yongshan Zhang}
\email{yszhang.cug@gmail.com}
\affiliation{
  \institution{China University of Geosciences Wuhan}
  \country{China}
}

\author{Yicong Zhou}
\email{yicongzhou@um.edu.mo}
\authornotemark[1]
\affiliation{
  \institution{University of Macau}
  \country{Macau, China}
}

\renewcommand{\shortauthors}{Trovato et al.}

\begin{abstract}

Identifying the land cover category for each pixel in a hyperspectral image (HSI) relies on spectral and spatial information. An HSI cuboid with a specific patch size is utilized to extract spatial-spectral feature representation for the central pixel. In this article, we investigate that scene-specific but not essential correlations may be recorded in an HSI cuboid. This additional information improves the model performance on existing HSI datasets and makes it hard to properly evaluate the ability of a model. We refer to this problem as the spatial overfitting issue and utilize strict experimental settings to avoid it. We further propose a multiview transformer for HSI classification, which consists of multiview principal component analysis (MPCA), spectral encoder-decoder (SED), and spatial-pooling tokenization transformer (SPTT). MPCA performs dimension reduction on an HSI via constructing spectral multiview observations and applying PCA on each view data to extract low-dimensional view representation. The combination of view representations, named multiview representation, is the dimension reduction output of the MPCA. To aggregate the multiview information, a fully-convolutional SED with a U-shape in spectral dimension is introduced to extract a multiview feature map. SPTT transforms the multiview features into tokens using the spatial-pooling tokenization strategy and learns robust and discriminative spatial-spectral features for land cover identification. Classification is conducted with a linear classifier. Experiments on three HSI datasets with rigid settings demonstrate the superiority of the proposed multiview transformer over the state-of-the-art methods.

\end{abstract}




\maketitle

\section{Introduction}
Hyperspectral images (HSIs) are composed of hundreds of spectral bands collected by remote sensors, which enable precise scene understanding \cite{hsi1,hsi2}. HSI classification aims at identifying land cover categories of pixels in HSIs \cite{hsi3}. It provides valuable information for various applications, including mineral exploration \cite{app1}, object detection \cite{app2} and precise agriculture \cite{app3}. Since building an HSI dataset for classification is complex and expensive, only a few datasets with limited labeled samples and categories are available. The number of samples in different classes varies significantly. Besides, noise may disrupt the characteristics of the spectrum. Therefore, designing a model for high-quality classification is challenging. Since scenes recorded in HSI datasets are much simpler than real-world scenarios, rigorous experimental settings should be used to assess the ability of a model.

Numerous HSI classification models have been proposed over the past few decades \cite{dl1, dl2}. Classical machine learning classifiers, such as support vector machine (SVM) \cite{svm} and K-nearest neighbors (KNN) \cite{knn}, are conducted for HSI classification. To enhance the output quality, hand-designed feature extraction methods are adopted to extract discriminative representations \cite{cfea1, cfea2}. However, it is difficult to design proper feature extraction algorithms \cite{hsi3}. The output quality of these methods is unsatisfactory. Deep learning (DL) models have advanced in the last decades, achieving better performance than traditional learning methods on many applications \cite{dlapp1,dlapp2}. DL-based HSI classification models are widely researched \cite{dl1,dl3}. Most DL-based models are composed of a feature extraction block and a classifier. Convolutional neural networks (CNNs) \cite{cnn} and transformer networks \cite{transformer} are widely adopted for feature extraction \cite{cnn1,hit}. To capture spatial and spectral correlations, the dual-branch structure is proposed to separately extract spatial and spectral features \cite{cnn2}. Recently, graph neural networks (GNN) are explored to model spatial and spectral correlations in HSIs \cite{gnn1, gnn2}. These DL-based models innovate in model design and greatly improve performance quality. However, the high dimensionality of HSIs may cause high computation cost and the `curse of dimensionality' problem in DL models \cite{curse}. Principal component analysis (PCA) or band selection is executed to reduce the dimension of HSI in some works \cite{ssftt}. But it may discard detailed spectral information. Besides, these works conduct feature extraction on HSI cuboids with large spatial patch sizes, providing additional scene-specific but not general spatial-spectral information for classification. Therefore, performance on test data cannot properly evaluate the ability of the model. Moreover, separate extracting spatial and spectral features in dual-branch may neglect the spatial and spectral correlations and cause model redundancy.

Multiview observations provide comprehensive descriptions of the same object from multiple perspectives, such as face images of a person in different light conditions \cite{multiview}. Multiview data improve understanding and enhance the performance of DL models. A multiview model is roughly composed of multiview construction, interaction and fusion \cite{dl3}. Essentially, an HSI is a multiview observation of a scene captured by hyperspectral imaging sensors, recording spectral information in values of bands. Existing methods construct multiview observations from HSIs at the band or feature level \cite{mv1}. The views in band-level multiview construction strategies are subsets of bands. Subset division is based on correlations between bands \cite{mv3} or random selection strategies \cite{mv2}. DL models are then applied to multiview data to aggregate information from each view and extract feature representations. Though subset splitting decomposes an HSI into multiple low-dimensional view observations, processing all views still requires high computation cost. On the other hand, since spatial and spectral characteristics are two critical clues for object identification in HSIs, most existing methods extract spatial and spectral features separately and construct multiview observations at the feature level \cite{mv4}. These features are combined for classification. Although the aforementioned approaches have been proven to be effective in HSI classification, adopting the multiview framework for HSI classification still remains an open question. 

In this paper, we investigate the spatial correlations in HSIs and discover the spatial overfitting issue in the existing methods. Scenes recorded in existing HSI datasets are relatively simple. Scene-specific information, such as relative location, may provide additional but not essential spatial information for classification. Therefore, under some experimental settings, performance on HSI datasets cannot properly indicate the ability of the models. Both CNN-based and transformer-based methods may suffer from this problem. We find that methods using an HSI cuboid with a large patch size are prone to encounter this problem. To avoid the above issue, we propose a multiview transformer with rigid experimental settings. Our proposed method is composed of multiview principal component analysis (MPCA), spectral encoder-decoder (SED) and spatial-pooling tokenization transformer (SPTT). MPCA conducts dimension reduction for an HSI. The multiview observations are constructed from an HSI based on the spectral correlation. PCA is applied to each view data to obtain a low-dimensional view representation. The combination of all view representations, named the multiview representation of an HSI, is the dimension-reduction output of the MPCA. The multiview representations are fused using a fully-convolutional SED. We further propose SPTT to extract discriminative spatial-spectral feature representations for classification. The multiview transformer is constructed using CNN-based blocks and transformer-based modules, taking advantage of both popular networks in feature extraction. Experiments demonstrate the proposed method can learn spectral and spatial characteristics in an HSI cuboid with a small patch size using a small number of training samples. The contributions of this paper are summarized as follows.

\begin{itemize}
    \item We explore the spatial correlation in an HSI cuboid with different patch sizes and pose the spatial overfitting issue in existing HSI classification methods. The scene-specific correlations in HSI datasets provide additional but not essential spatial-spectral information for classification. 

    \item To reduce the spectral dimension of an HSI while preserving details, we propose a multiview framework for dimension reduction and information fusion. The multiview principal component analysis is used to extract low-dimensional multiview representations from an HSI. Multiview information is then aggregated via spatial encoder-decoder to obtain multiview features.

    \item We develop a spatial-pooling tokenization transformer for HSI feature extraction. Spatial-pooling tokenization strategy transforms the multiview features into tokens. Each token represents local information in a region. Scene-specific correlations are smoothed in this process. The multi-head attention mechanic is utilized to learn robust and discriminative spatial-spectral features for classification. 
\end{itemize}

The rest of this article is organized as follows: \cref{sec:spatial_overfitting} explores the spatial overfitting issue in the existing DL-based HSI classification methods. \cref{sec:method} introduces the proposed multiview transformer for HSI classification. Experimental results and model analysis are presented in \cref{sec:res}. Finally, \cref{sec:conclusion} draws the conclusion.


\begin{figure}[t]
  \centering
   \includegraphics[width=0.75\linewidth]{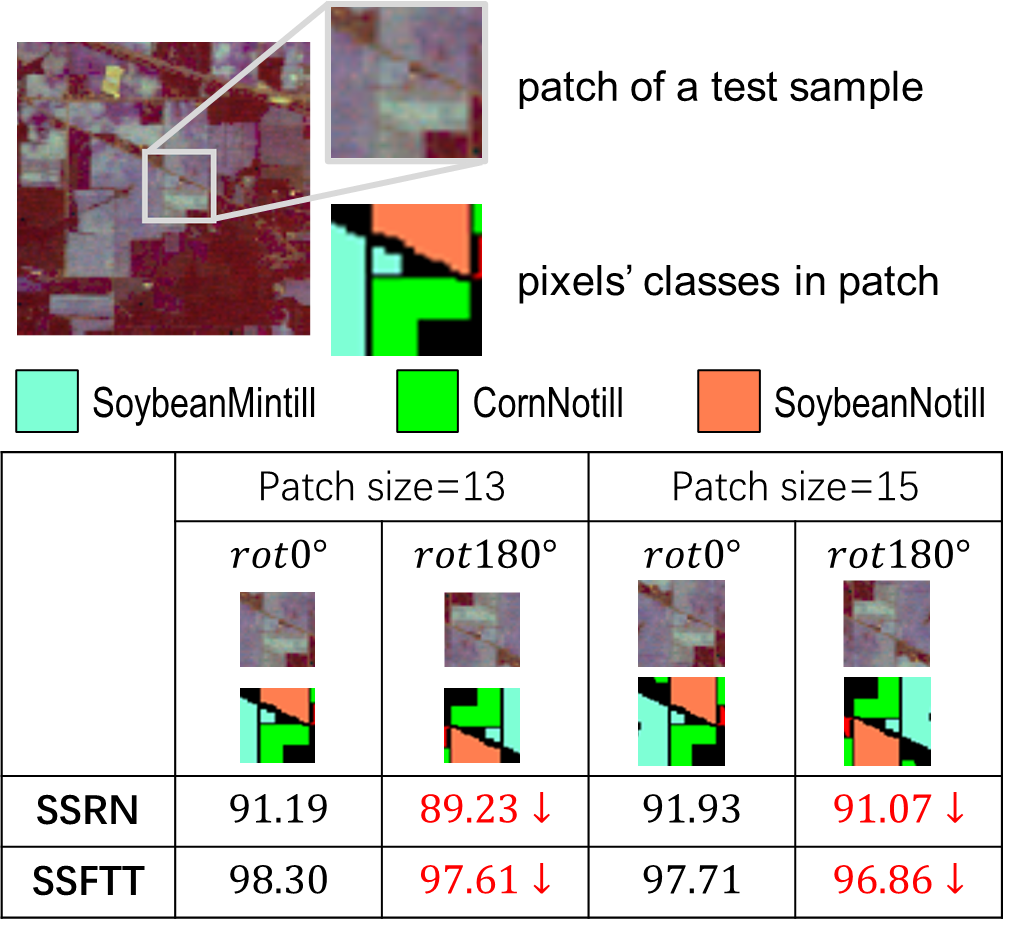}
   \caption{Quantitative analysis of the spatial overfitting issue in HSI classification methods.}
   \label{fig:spatial_overfitting}
\end{figure}

\begin{figure*}[t]
  \centering
   \includegraphics[width=0.95\linewidth]{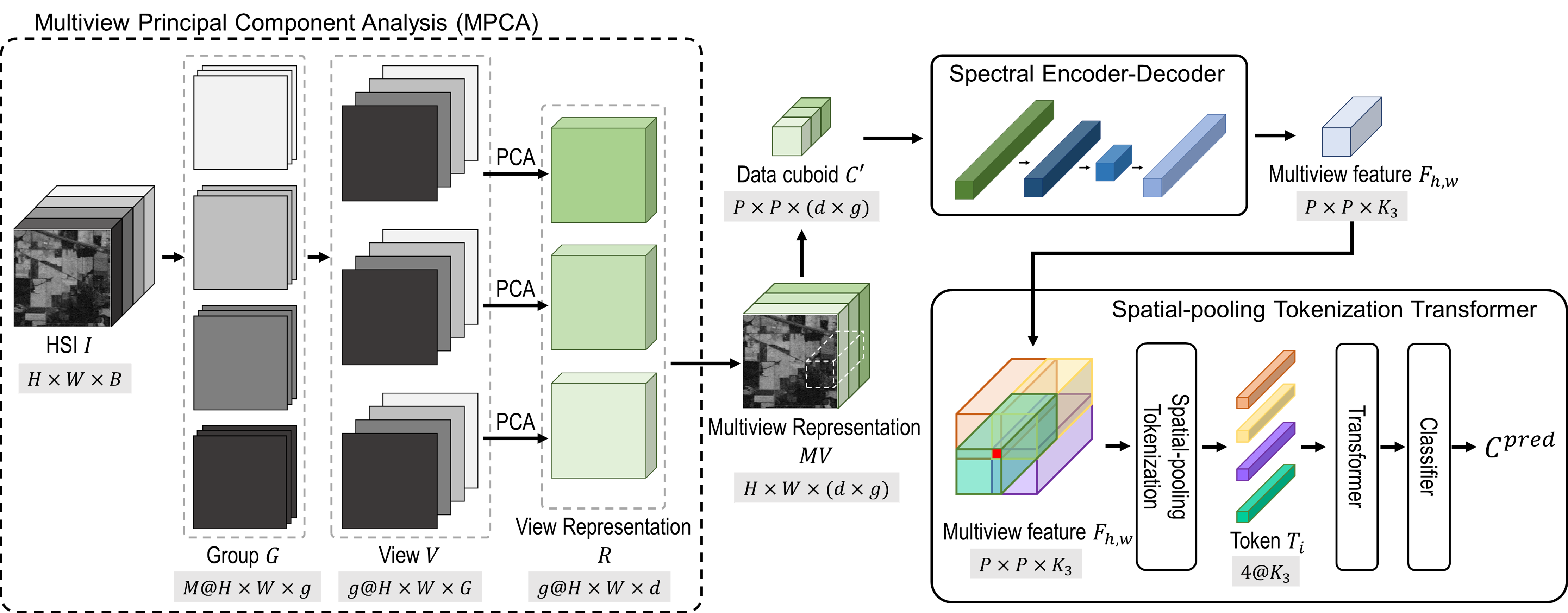}
   \caption{Overview illustration of the proposed multiview transformer. The multiview transformer consists of multiview principal component analysis (MPCA), spectral encoder-decoder (SED) and spatial-pooling tokenization transformer (SPTT).}
   \label{fig:model}
\end{figure*}

\section{Spatial Overfitting Issue}
\label{sec:spatial_overfitting}

Given a hyperspectral image $I$ with spatial size $H \times W$ and $B$ bands, data cuboid $C_{h,w} \in \mathcal{R}^{P \times P \times B}$ centered at spatial location $(h,w)$ with $P \times P$ surrounding pixels is captured to extract spatial-spectral features for identifying the land cover type at $(h,w)$. $I_{h,w,b}$ represents an element at location $(h,w)$ and $b$-th band. Neighboring spectrums $I_{h',w',:} \in C_{h,w}$ provide spatial context and spectral information for feature extraction. Based on the properties of HSIs, we investigate the spatial information in a data cuboid $C_{h,w}$. Objects recorded in neighboring pixels $I_{h',w',:}$ provide important information for land cover identification of $I_{h,w,:}$ since $I_{h',w',:}$ and $I_{h,w,:}$ may be highly correlated. On the other hand, items at a distance may be unrelated to $I_{w,h,:}$. These pixels may contain invalid information for robust feature extraction and increase the computational cost. Moreover, these redundant pixels may be noise that damages the general spatial and spectral features. Therefore, the patch size $P$ of the data cuboid $C_{h,w}$ should be carefully decided according to the properties of an HSI. However, existing DL-based models are evaluated on different settings of $P$, regardless of the geographical relations between pixels \cite{superpca}. Some methods achieve the best performance using a large $P$, such as $P=13$ \cite{ssftt}. Optimal $P$ for the same dataset is varying in different models. An HSI cuboid $C_{h,w}$ with a large patch size $P$ contains pixels far away from $I_{h,w,:}$. Considering the spatial resolutions of HSI datasets for land-cover classification ($1-20m/pixel$), it indicates that information in long-distance pixels improves the understanding of $I_{h,w,:}$, which goes against our intuitive understanding. Besides, unsatisfying outputs are obtained by these models when using small values of $P$. 

We investigate the causes of the experimental observations. Using a small patch size $P$, neighboring elements $I_{h',w',:}$ provide related spatial-spectral information for the classification of $I_{h,w,:}$. At this stage, the performance improves when increasing the patch size $P$. The output quality is supposed to become worse if we use a large patch size since noisy information is included in the HSI cuboid $C_{h,w}$. However, scenes recorded in the existing HSI datasets are quite simple. Objects in these scenes have specific distribution patterns of location. These scene-specific spatial correlations are more likely to be recorded in data cuboids with large patch sizes, providing additional but not general features for classification. We further conduct experiments on rotated test samples to verify our assumptions \cite{rian}. As shown in \cref{fig:spatial_overfitting}, we present results obtained by CNN-based SSRN \cite{ssrn} and transformer-based SSFTT \cite{ssftt}. The rotated test samples change the relative location in training data, achieving lower accuracies. It implies that existing models can learn scene-specific spatial-spectral information from training data cuboids with large patch sizes. 

We denote the above observation as the spatial overfitting issue, illustrating that existing models learn scene-specific but not general spatial and spectral correlations from training data. Since only a few HSI datasets are accessible, to avoid the spatial overfitting issue and properly evaluate the performance of a model, rigorous experimental settings should be used. The model should be carefully designed to extract robust and discriminative features for classification. Moreover, large-scale HSI datasets for classification are required in the big data era \cite{bigdata1,bigdata2}.



\section{Method}
\label{sec:method}

To avoid the spatial overfitting issue, we propose the multiview transformer framework for HSI classification. This section presents the details of the multiview transformer. We first describe the multiview principal component analysis (MPCA) for HSI dimension reduction. Then, we introduce the spectral encoder-decoder (SED) to aggregate multiview information. Finally, we propose the spatial-pooling tokenization transformer for spatial-spectral feature extraction and classification. The overall framework of the proposed multiview transformer is shown in \cref{fig:model}

\subsection{Multiview Construction \& Representation}

Compared to a color image, a hyperspectral image has rich spectral information stored in bands, providing details for land cover identification. However, there is redundancy in adjacent bands \cite{bandselect}. Besides, high-dimensional input data may cause the `curse of dimensionality' \cite{curse} and high computation cost. Therefore, some existing methods adopt PCA to reduce the dimensionality of the original HSI. The computational complexity of PCA rapidly grows with respect to the number of bands. Applying PCA to the original data may discard detailed information. We introduce the multiview PCA for HSI dimension reduction. Given a hyperspectral image $I \in \mathcal{R}^{H \times W \times B}$ with $H \times W$ spatial size and $B$ spectral bands, we first conduct min-max scaling normalization (MMNorm) to rescale element values in $I$ to $\left [ 0, 1 \right ] $ as follow:
\begin{equation}
    I' = \frac{I - \min I}{\max I-\min I} .
\end{equation}

\noindent Each term $I_{h,w,b}$ in the HSI is normalized using the global maximum and minimum values of $I$. Band-wise normalization strategies are widely adopted in the existing methods \cite{rian}. The value range differs in different bands for the existing HSI datasets. We assume that the maximum and minimum values that can be reached in all bands are the same. However, the expressibility of bands is not fully leveraged for existing datasets since scenes recorded in HSIs are simple. Band-wise normalization methods may distort the spectrums. MMNorm aims at corrupting the effect of the value range in spectral feature extraction while preserving the spectral characteristics. We split the $I'$ into $M$ groups $\{G_m\}_{m=1}^{M}$. Each group contains $g = \left \lceil B/M \right \rceil$ continuous bands as follows:
\begin{equation}
    G_{m}=[I'_{:,:,m \times g}, I'_{:,:,m \times g + 1}, ...,I'_{:,:,(m+1) \times g -1}]. 
\end{equation}

\noindent The $I'$ is padded with zeros in the spectral dimension for group construction. Since a band and its neighboring bands are linearly correlated \cite{bandselect}, bands in each group $G_{m}$ provide similar observations to the scene. Different groups store varying spectral information. All groups construct the complete observation for classification. Direct extracting low-dimensional representation from each group may cause biased observation and ignorance of global correlations in bands. Therefore, we further construct multiview observations $\{V_n\}_{n=1}^{g}$ from groups $G$ using: 
\begin{equation}
    V_{n} = [G^{n}_{1}, G^{n}_{2},\dots,G^{n}_{M} ],
\end{equation}

\noindent where $G_{m}^{n}$ is the $n$-th band in $G_{m}$. Each view contains a non-repeating band $G_{m}^{n}$ in each group. Using this multiview construction strategy, each view contains comprehensive spectral information about the scene. We perform PCA on each view to extract low-dimensional view representation $\{R_{n}\}_{n=1}^{g}$. The high-dimensional spectral vector of each pixel in view $V_{n}$ is mapped to feature space with dimension $D_{n}$. In this paper, we use the same $D_{n}=d$ for all views. All view representations are then concatenated in the spectral dimension to get multiview representation $M \in \mathcal{R}^{H \times W \times (g \times d)}$ as follows:
\begin{equation}
    MV=[R_{1}, R_{2}, \dots, R_{g}].
\end{equation}

\noindent The multiview representation $MV$ is the dimension reduction output of the original HSI. Important global spectral details are highlighted in each view representation $R_{n}$. All view representations provide complementary observations to each other. Besides, the dimensionality of each view data $V_{n}$ is much smaller than the original HSI, which can reduce the complexity of applying PCA.

\subsection{Spectral Encoder-Decoder}
Let $C'_{h,w} \in \mathcal{R}^{P \times P \times (g \times d)}$ be the multiview cuboid cropped from $MV$ centered at $(h,w)$. Motivated by encoder-decoder networks \cite{unet}, we conduct a $3$-layer fully convolutional block, named SED, to aggregate multiview spatial and spectral information and extract multiview feature maps $F_{h,w}$. A cuboid $C'_{h,w}$ with a small patch size $P$ is used to avoid spatial overfitting issues. The first layer in the SED is a CNN with $K_{1}$ three-dimensional kernel with shape $k^{1}_{1} \times k^{1}_{2} \times k^{1}_{3}$. We adopt the 3D-CNN to adaptively aggregate multiview representation. The correlation of multiview spectral representation is implicitly explored and fused in a heuristic way using 3D kernels. Outputs of each kernel are combined to get a feature cuboid $F^{1}_{h,w}$. The feature cuboid is then fed into 2D convolutional layers with kernel size $k^{2}_{1} \times k^{2}_{2}$ and $k^{3}_{1} \times k^{3}_{2}$, sequentially. The numbers of kernels in these layers are $K_{2}$ and $K_{3}$, respectively. In our settings, the U-shape structure in encoder-decoder means $(K_{1} \times k^{1}_{3} ) > K_{2}$ and $K_{3} > K_{2}$. We conduct the fully convolutional layers to extract spatial-spectral feature maps $F_{h,w}$ for multiview representation $C'_{h,w}$. We pad each layer's inputs with zeros to maintain the spatial size $P \times P$ between inputs and outputs, preserving the pixel correlations in $F_{h,w}$. The U-shape framework in the spectral dimension first encodes the multiview representation of pixels into low-dimensional space and then maps to discriminative feature space. 

\begin{figure}[t]
  \centering
   \includegraphics[width=0.99\linewidth]{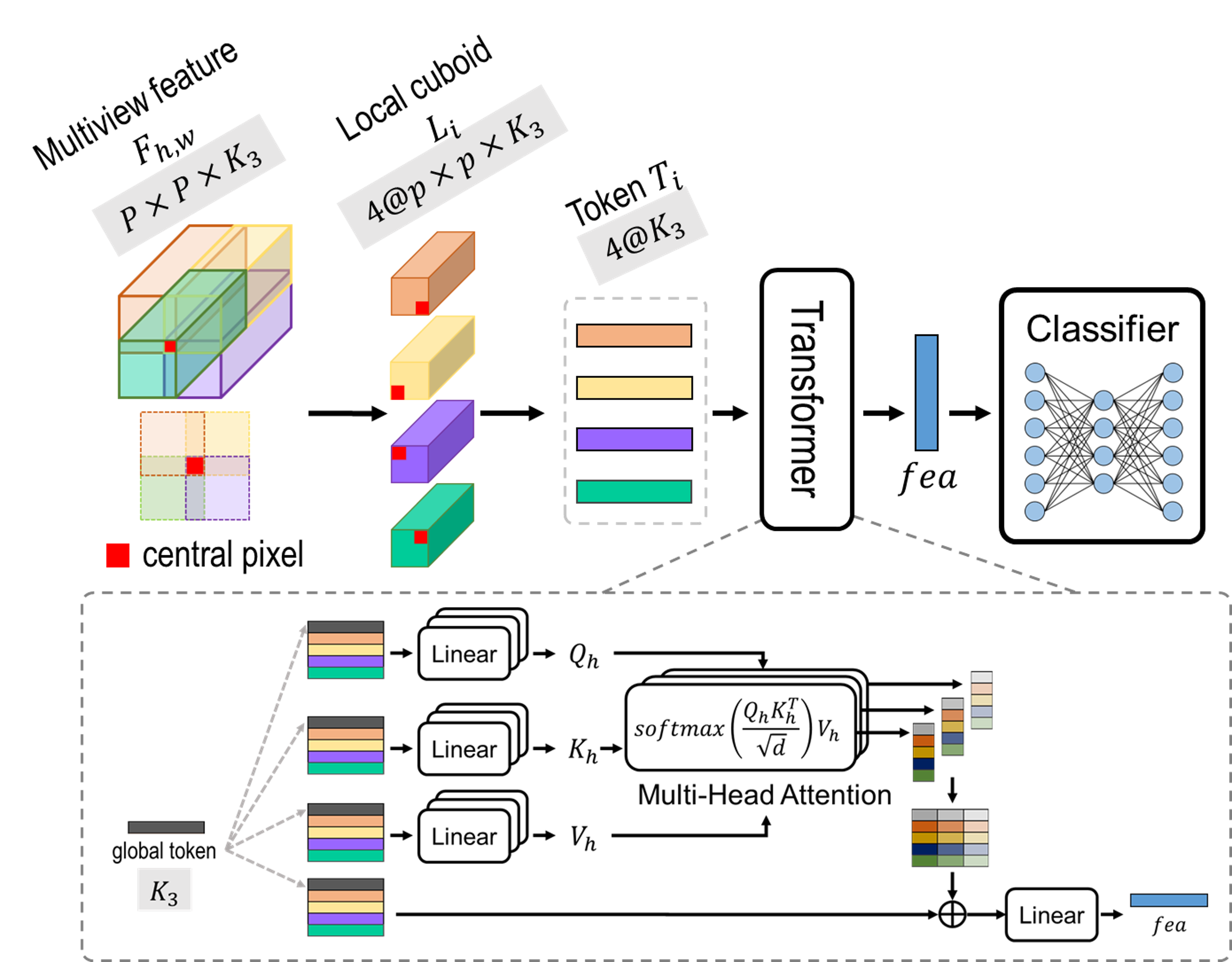}
   \caption{Illustration of the spatial-pooling tokenization transformer.}
   \label{fig:sptt}
\end{figure}

\subsection{Spatial-Pooling Tokenization Transformer}
We get a multiview feature cuboid $F_{h,w} \in \mathcal{R}^{P \times P \times K_{3}}$ from SED. Spatial-pooling tokenization transformer (SPTT) is adopted to extract discriminative features for classification. Pixel-wise and band-wise tokenization methods are investigated in the existing transformer-based methods \cite{sf,ssftt}. These strategies generate a large number of tokens, resulting in the high computational complexity of attention modules \cite{transformer}. Besides, since transformer networks are proposed for natural language processing tasks, not all modules are required when designing transformer-based HSI processing models. We design a spatial-pooling tokenization transformer, which is suitable for HSI feature extraction. The spatial-pooling tokenization method generate only four tokens $\{T_i\}_{i=1}^{4}$ from a multiview feature cuboid $F_{h,w}$. Each token $T_{i}$ has spectral and spatial information in a specific geographical direction in $F_{h,w}$. Since the location information is implicitly encoded in the token, position encoding in the transformer networks is not required. A parameterized global token is appended to learn global information of an HSI in the training process and give additional clues for inference. Multi-head attention modules learn robust and discriminative features and classification is conducted using a linear classifier. 

\textbf{Spatial-Pooling Tokenization.} As shown in \cref{fig:sptt}, we first decompose the feature cuboid $F_{h,w}$ into four small local cuboids $\{L_{i}\}_{i=1}^{4}$ with shape $p \times p \times K_{3}$. Each local cuboid $L_{i}$ contains the center pixel since it includes the most important information for classification at location $(h,w)$. These four cuboids provide spatial-spectral features in four directions in $F_{h,w}$, respectively. The average values of bands in each local cuboid $L_{i}$ are used as the token representation $T_{i}$:
\begin{equation}
    T_{i}=[\frac{\sum{L^{1}_{i}}}{ p \times p}, \frac{\sum{L^{2}_{i}}}{ p \times p}, \dots , \frac{\sum{L^{K_{3}}_{i}}}{ p \times p}],
\end{equation}

\noindent where $L^{c}_{i}$ is the feature map of $L_{i}$ in $c$-th channel. The spatial-pooling tokenization is a parameter-free method to generate only four vector tokens $T_{i}$ from $F_{h,w}$. Besides, scene-specific information may be smoothed using the spatial-pooling strategy. This may help to avoid the spatial overfitting issue. 


\textbf{Transformer for Feature Extraction.} Attention mechanic in transformer 
 networks draws global dependencies between inputs and outputs, capturing the correlations between input tokens. We adopt the $H$-head attention to extract robust spatial-spectral features for classification. A parameterized global token $T_{global} \in \mathcal{R}^{K_{3}}$ is append at the first position as follows:
\begin{equation}
    Tokens=[T_{global}, T_{1}, T_{2}, T_{3}, T_{4}].
\end{equation}

\noindent The global token $T_{global}$ is updated in the training process, learning the global information from the training data. It provides additional clues for interference. In transformer networks, positional encoding indicates the location correlation of phrases in a sentence or patches in an image. We remove the positional encoding since the token index $i$ implies the region of feature encoded in $T_{i}$. Additional position information is not required. In single-head attention, three learnable weight matrices, $W^{h}_{Q}$, $W^{h}_{K}$ and $W^{h}_{V}$, linearly map $Tokens$ to queries $Q_{h}$, keys $K_{h}$ and values $V_{h}$, respectively. The dimension of queries, keys and values is $d$. The superscript $h$ of $W^{h}_{Q}$, $W^{h}_{K}$ and $W^{h}_{V}$ denotes the number of heads in $H$-head attention. We reserve it here for future use. The formulation of the attention module is given by:
\begin{equation}
    Att_{h}=softmax(\frac{Q_{h}K^{T}_{h}}{\sqrt{d } } )V_{h}.
\end{equation}

\noindent The $H$-head attention employs $H$ attention layers in parallel to extract rich and varying features. As shown in \cref{fig:sptt}, outputs of $H$-head attention layers are concatenated as:
\begin{equation}
    A=[Att_{1}, Att_{2}, \dots, Att_{H}].
\end{equation}

\noindent We adopt the residual connection \cite{resnet} to fuse multi-level features and speed up the training process as follows:
\begin{equation}
    TA=A+T.
\end{equation}

\noindent To conduct element-wise addition on $A$ and $T$, the dimension $d$ is determinded by:
\begin{equation}
\label{eqa:khd}
    d=\frac{K_{3}}{H},
\end{equation}

\noindent where $d$ is the same in each attention block. We further use a linear layer to fuse output features and extract discriminative spatial-spectral features $fea \in \mathcal{R}^{j}$ for classification:
\begin{equation}
    fea=Linear(TA).
\end{equation}

\textbf{Classification.} The output feature vectors $fea$ learned by the SPTT are fed into a classifier. We adopt a fully-connected layer with a softmax function as a classifier, which can be formulated by
\begin{equation}
    C^{pred}=softmax(FC(fea)),
\end{equation}

\noindent where $FC$ denotes the fully connected layer. 

We use CNN-based and transformer-based blocks to construct our model, leveraging the strengths of these blocks for HSI feature extraction. Our proposed framework aggregates spatial and spectral information in the same branch, reducing the redundancy of the model. Moreover, we carefully design the network structure to extract robust and general spatial-spectral features while avoiding the spatial overfitting issue.

\begin{figure*}
  \centering
    \begin{subfigure}{0.12\linewidth}
    \includegraphics[width=1\linewidth]{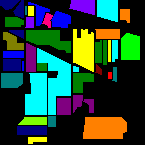}
    \caption{Ground truth}
  \end{subfigure}
  \hfill
    \begin{subfigure}{0.12\linewidth}
    \includegraphics[width=1\linewidth]{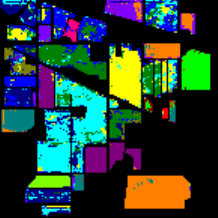}
    \caption{SSAN}
  \end{subfigure}
  \hfill
    \begin{subfigure}{0.12\linewidth}
    \includegraphics[width=1\linewidth]{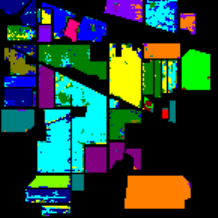}
    \caption{SSRN}
  \end{subfigure}
  \hfill
  \begin{subfigure}{0.12\linewidth}
    \includegraphics[width=1\linewidth]{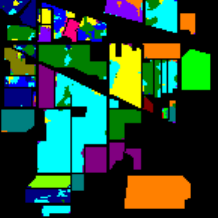}
    \caption{RIAN}
  \end{subfigure}
  \hfill
    \begin{subfigure}{0.12\linewidth}
    \includegraphics[width=1\linewidth]{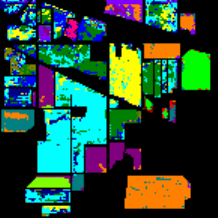}
    \caption{SF}
  \end{subfigure}
  \hfill
  \begin{subfigure}{0.12\linewidth}
    \includegraphics[width=1\linewidth]{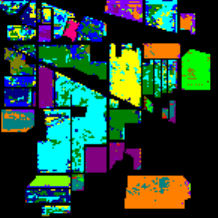}
    \caption{HiT}
  \end{subfigure}
    \hfill
  \begin{subfigure}{0.12\linewidth}
    \includegraphics[width=1\linewidth]{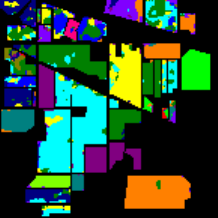}
    \caption{SSFTT}
  \end{subfigure}
    \hfill
  \begin{subfigure}{0.12\linewidth}
    \includegraphics[width=1\linewidth]{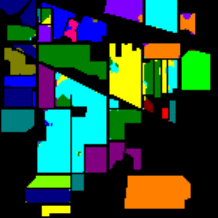}
    \caption{Ours}
  \end{subfigure}
  \caption{Ground truth and classification maps obtained by different methods on the IP dataset. }
  \label{fig:ipvis}
\end{figure*}

\subsection{Implementation}

We first construct 10 views from a scaled HSI. PCA is applied to each view in the spectral dimension. In each view, three principal components are preserved as a view representation for each pixel. We concatenate all view representations to get a multiview representation cuboid with shape $P \times P \times 30$. 

In the first layer of SED, 3-D convolution with eight $3 \times 3 \times 3$ kernels is adopted to aggregate multiview information. We utilize the zero-padding strategy to maintain the shape of the input and output of each kernel. Outputs of all kernels are combined to generate a feature cuboid with shape $P \times P \times 240$. Then, $3 \times 3$ convolutional layers with kernels size 40 and 64 are employed to feature cuboid sequentially. The output shape of the SED is $P \times P \times 64$. The spatial size is preserved in all feature cuboids in SED for retaining spatial correlation while aggregating multiview spectral information.

Spatial-pooling tokenization method generates four token cuboids with shape $\lceil P/2 \rceil \times \lceil P/2 \rceil \times 64$. The average values of each channel are calculated to generate a vector token with a length of 64 from each token cuboid. Tokens are then concatenated and a parameterized global token vector is appended at the first position to obtain spatial-spectral tokens $Tokens \in  \mathcal{R}^{5 \times 64}$. The dimension of queries, keys and values are set to $d=8$. We utilize $8$-head attention to learn comprehensive and robust feature representations. The outputs of each attention module are concatenated and added to the input token. A fully-connected layer is adopted to obtain the final spatial-spectral representations. A linear layer with a softmax function is utilized as a classifier in our implementation.


\section{Experiments}
\label{sec:res}

\subsection{Data Description}
In this section, three public hyperspectral classification datasets, Indian Pines (IP), Pavia University (PU) and Houston 2013 (Houston) datasets, are employed to evaluate our proposed multiview transformer. The Indian Pines dataset was captured by the airborne visible/infrared imaging spectrometer (AVIRIS) sensor, recording a scene from Northwest Indiana. It consists of $145 \times 145$ pixels with a spatial resolution of $20m$. Each pixel is described by 200 spectral bands. There are 16 different land-cover classes recorded in the IP dataset. The PU dataset was acquired by the reflective optics system imaging spectrometer (ROSIS) sensor over Pavia University. The size of the PU dataset is $610 \times 340$ pixels and 103 bands, with a spatial resolution of $1.3m$ per pixel. The ground truth of the PU dataset contains 9 land-cover categories in the scene. Houston database was collected by the ITRES-CASI 1500 sensor over the University of Houston and its surrounding area. The Houston dataset was accessible by the 2013 IEEE GRSS Data Fusion Contest. It includes $349 \times 1905$ pixels with a spatial resolution of $2.5m$ and 144 spectral bands. There are 15 different types of land covers recorded in this scene. 

\begin{table}
\caption{Classification performance of different methods on the IP dataset.}
    \label{tab:res_ip}
    \centering
    \begin{tabular}{c|cc|cc|cc}
    \toprule
          \multirow{2}*{Methods}& \multicolumn{2}{c}{patch size=3} \vline &\multicolumn{2}{c}{patch size=5} \vline &\multicolumn{2}{c}{patch size=7} \\ \cline{2-7}
         ~ & OA & AA  &OA  &AA  &OA  &AA \\ \hline
         RSSAN & 64.81 & 51.73 & 66.60 & 56.47 & 68.30 & 61.89 \\
         SSRN & 68.87 & 52.17 & 86.53 & 74.53 & 91.28 & 79.10 \\
         RIAN & 78.33 & 74.46 & 86.35 & 90.19 & 89.27 & 89.97 \\
         SF   & 58.94 & 44.39 & 62.94 & 45.48 & 66.58 & 54.00 \\
         HiT  & 68.78 & 54.88 & 71.26 & 62.99 & 71.01 & 56.11 \\
         SSFTT& -     & -     & 71.51 & 48.03 & 86.31 & 68.19 \\ \hline
         Ours & \textbf{88.01} & \textbf{87.63} & \textbf{94.14} & \textbf{95.34} & \textbf{96.83} & \textbf{96.92} \\
         \bottomrule
    \end{tabular}
\end{table}

\begin{table}
\caption{Classification performance of different methods on the PU dataset.}
    \label{tab:res_pu}
    \centering
    \begin{tabular}{c|cc|cc|cc}
    \toprule
          \multirow{2}*{Methods}& \multicolumn{2}{c}{patch size=3} \vline &\multicolumn{2}{c}{patch size=5} \vline &\multicolumn{2}{c}{patch size=7} \\ \cline{2-7}
         ~ & OA & AA  &OA  &AA  &OA  &AA \\ \hline
         RSSAN & 92.78 & 91.24 & 93.73 & 91.06 & 94.12 & 90.99 \\
         SSRN & 95.07 & 93.48 & 97.89 & 96.96 & 98.41 & 97.18\\
         RIAN & 94.08 & 89.15 & 97.34 & 97.14 & 98.83 & 98.08 \\
         SF   & 78.37 & 64.09 & 83.09 & 70.19 & 84.45 & 67.44 \\
         HiT  & 95.33 & 93.60 & 96.13 & 94.71 & 96.92 & 94.70 \\
         SSFTT& - & - & 93.17 & 90.83 & 97.79 & 96.14 \\ \hline
         Ours & \textbf{97.18} & \textbf{96.20} & \textbf{98.98} & \textbf{98.34} & \textbf{99.65} & \textbf{99.35} \\
         \bottomrule
    \end{tabular}
\end{table}

\subsection{Experimental Setup}
\indent \textbf{Compared Methods.} To verify the superiority of the proposed method, we select several advanced CNN-based and transformer-based methods for comparison. The CNN-based methods chosen in this paper are residual spectral-spatial attention network (RSSAN) \cite{rssan}, spectral–spatial residual network (SSRN) \cite{ssrn} and rotation-invariant attention network (RIAN) \cite{rian}. As for transformer-based methods, there are patch-wise SpectralFormer (SF) \cite{sf}, hyperspectral image transformer (HiT) \cite{hit} and spectral-spatial feature tokenization transformer (SSFTT) \cite{ssftt}.

\noindent \textbf{Implementation Details.} The experiments of the proposed method and compared methods are conducted with PyTorch \cite{pytorch} or TensorFlow \cite{tensorflow} on NVIDIA 3090 GPU. To investigate the models' performance, we conduct experiments using different patch sizes, including 3, 5 and 7. We randomly select $5\%$, $5\%$ and $90\%$ labeled samples for model training, validation and testing, respectively. Each category has at least one training sample. Small patch sizes and a small amount of training samples are utilized to avoid the spatial overfitting issue. The batch size is set to 64. Each model has 300 training epochs applied. Parameters in each network are updated using Adam optimizer \cite{adam} with a learning rate $1e^{-4}$ in each iteration. The overall accuracy (OA) and average accuracy (AA) are introduced for quantitative analysis and comparison. Different from OA, AA provides an informative evaluation of the model performance by considering the classification accuracy of each class, which is an important indicator of output quality on class-imbalanced HSI datasets.

\begin{table}
\caption{Classification performance of different methods on the Houston dataset.}
    \label{tab:res_grss}
    \centering
    \begin{tabular}{c|cc|cc|cc}
    \toprule
          \multirow{2}*{Methods}& \multicolumn{2}{c}{patch size=3} \vline &\multicolumn{2}{c}{patch size=5} \vline &\multicolumn{2}{c}{patch size=7} \\ \cline{2-7}
         ~ & OA & AA  &OA  &AA  &OA  &AA \\ \hline
         RSSAN & 83.77 & 82.17 & 84.70 & 83.21 & 84.05 & 81.75 \\
         SSRN & 80.28 & 79.59 & 85.20 & 84.41 & 90.55 & 89.72 \\
         RIAN & 91.15 & 91.38 & 93.14 & 92.69 & 93.52 & 93.63 \\
         SF   & 77.27 & 79.54 & 77.71 & 78.90 & 78.15 & 79.35 \\
         HiT  & 85.56 & 85.33 & 86.94 & 86.25 & 89.00 & 88.08 \\
         SSFTT& - & - & 89.16 & 89.10 & 96.37 & 96.58 \\ \hline
         Ours & \textbf{96.04} & \textbf{96.56} & \textbf{98.21} & \textbf{98.40} & \textbf{98.56} & \textbf{98.57} \\
         \bottomrule
    \end{tabular}
\end{table}

\subsection{Experimental Results}
\cref{tab:res_ip}, \cref{tab:res_pu} and \cref{tab:res_grss} report the classification scores obtained by different methods for IP, PU and Houston, respectively. Visualization results of classification maps on the IP dataset with the patch size $P=7$ are shown in \cref{fig:ipvis}. The proposed multiview transformer achieves the best OA and AA on all datasets regardless of the patch size settings. The performance of existing methods becomes less reliable under settings of small patch sizes. The multiview transformer significantly improves OA and AA compared to existing methods. Specifically, $9.68\%$ OA and $13.17\%$ AA enhancements are obtained on the IP when patch size is $P=3$. While the IP and PU datasets have a serious class-imbalanced problem, our method is able to yield satisfactory classification results in each class when patch size is $P=3$ ($87.63\%$ AA on IP, $96.56\%$ AA on PU). The performance of the multiview transformer boosts when we increase the patch size to $P=5$ and $P=7$. It indicates that our method learns discriminative features using spatial and spectral information. We also evaluate our proposed method on a challenging dataset, Houston. The obtained competitive results show the ability of our method in complex scene understanding.

\subsection{Ablation Studies}
In our proposed multiview transformer framework, we design multiple modules, including MPCA, SED and SPTT, to extract robust feature representation. To better understand each component, we investigate them via ablation studies. Experiments in this section are conducted with patch size $P=5$. Experimental results are presented in \cref{tab:abla}.

\begin{table}
\caption{Abalation analysis of different modules in the proposed multiview transformer.}
\label{tab:abla}
\begin{tabular}{c | c c c|c c c}
    \toprule
    \multirow{2}*{Settings} & \multicolumn{3}{c|}{Components}  &  \multicolumn{3}{c}{OA} \\ \cline{2-7}
      ~& MPCA & SED & GT & IP & Houston & PU\\ \hline
     S1 & \tikzxmark & \tikzcmark & \tikzcmark& 94.19 & 98.12 & 99.47\\
     S2 & \tikzcmark & \tikzxmark & \tikzcmark & 91.82 & 94.39 & 98.91\\
     S3 & \tikzcmark & \tikzcmark& \tikzxmark & 96.45 & 98.14 & 99.55\\ \hline
     Ours & \tikzcmark & \tikzcmark &  \tikzcmark  & \textbf{96.68} & \textbf{98.56} & \textbf{99.65} \\
  \bottomrule
\end{tabular}
\end{table}

\begin{figure}[h]
  \centering
  \includegraphics[width=0.95\linewidth]{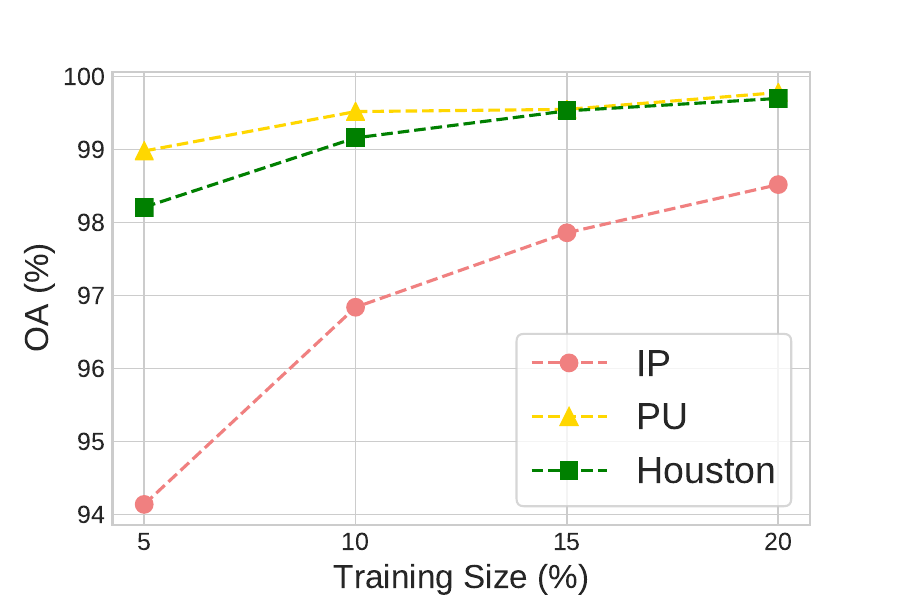}
  \caption{Performance analysis on three HSI datasets with different training sizes.}
  \label{fig:train_size}
\end{figure}

\textbf{The effectiveness of multiview PCA.} We adopt the MPCA to reduce the dimension of an HSI while preserving more spectral details. The classification accuracies on three HSI datasets decline if we replace the multiview PCA with the original PCA using the same output shape $H\times W \times 30$. The complexity of the PCA rapidly increases with input dimensionality. The multiview PCA is efficient since the dimension of each view data is much lower.

\textbf{Spectral encoder-decoder for multiview fusion.} The CNN-based SED is utilized to fuse multiview representations and extract multiview features. The encoder-decoder structure in spectral dimension learns the compact and expressive representation for each pixel in an HSI cuboid. We remove the SED and adopt a CNN layer to obtain an output with the same size. Experimental results indicate the importance of SED in multiview aggregation and feature extraction in our methods.

\textbf{Global token for feature extraction.} The parameterized global token (GT) learns the global information of an HSI in the training process, providing additional clues for interference. We notice that classification accuracy decreases when a zero vector is utilized as a global token. It implies that the global token contains general information about an HSI, which is effective for producing high-quality output.

\begin{figure}[h]
  \centering
  \includegraphics[width=0.95\linewidth]{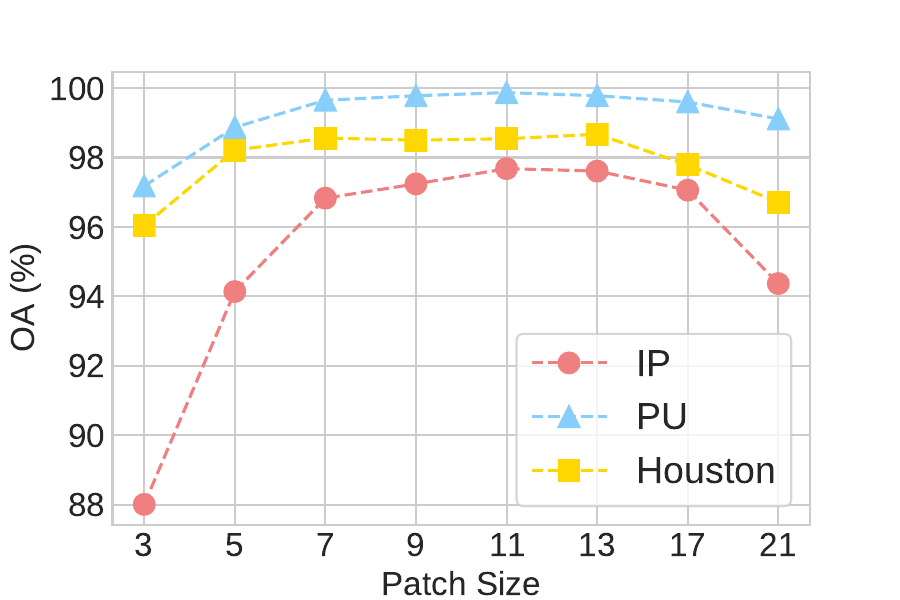}
  \caption{Performance analysis on three HSI datasets with different patch sizes.}
  \label{fig:patch_size}
\end{figure}

\subsection{Parameter Analysis}
The classification quality of the multiview transformer depends on the setting of each module. It is important to investigate the performance changes obtained by adjusting hyperparameters in our method. For this purpose, we conduct experiments for parameter sensitivity analysis. The OA changing trends on three HSI datasets are reported with different hyperparameter settings. The patch size is set to $P=5$ if not specified.

\begin{table}
\caption{Sensitivity analysis of the number of views $g$ in multiview construction.}
    \label{tab:abla_nview}
    \centering
    \begin{tabular}{c|ccccc}
    \toprule
        Datasets   & 5 & 8 & 10 & 12 & 15 \\ \hline
        IP      & 90.26 & 93.00 & 94.14 & 92.87 & \textbf{94.31} \\ \hline
        PU      & 98.92 & \textbf{99.09} & 98.98 & 98.77 & 98.76 \\ \hline
        Houston & 97.21 & 97.93 & \textbf{98.21} & 97.95 & 97.61 \\ 
    \bottomrule
    \end{tabular}
\end{table}

\textbf{Patch size for HSI cuboid.} The spectrums of adjacent pixels provide important spatial and spectral information for land cover identification. In our experimental settings, we adopt small patch sizes to avoid spatial overfitting. In \cref{fig:patch_size}, we further present classification scores obtained by using varying patch sizes to investigate the effectiveness of patch size for classification. Our method achieves competitive performance using small patch sizes. OAs continue to improve with the increase of patch sizes from $P=7$ to $P=11$ on the IP and PU datasets. While a data cuboid with a large patch size may contain unrelated noisy pixels, the scene-specific spatial information in the data cuboid and the adaptability of DL models make it possible to obtain a satisfactory output (OAs with $P=13$). The model may encounter the spatial overfitting problem in this process. However, 
we observe the declining changes with the continuous increase of patch size (OAs with $P=17$ and $P=21$). It can be explained that important spatial and spectral information may be buried by noise carried by a great number of unrelated pixels in the cuboid.


\begin{table}
\caption{Sensitivity analysis of the dimension $d$ of each view representation.}
    \label{tab:abla_ncom}
    \centering
    \begin{tabular}{c|ccccc}
    \toprule
         Datasets & 1 & 2 & 3 & 4 & 5 \\ \hline
        IP      & 83.79 & 90.40 & 94.14 & 94.21 & \textbf{94.57} \\ \hline
        PU      & 97.70 & 98.39 & \textbf{98.98} & 98.81 & 98.74 \\ \hline
        Houston & 92.35 & 96.95 & 98.21 & \textbf{98.45} & 97.95 \\ 
    \bottomrule
    \end{tabular}
\end{table}

\begin{table}
\caption{Sensitivity analysis of the number of heads $H$ and the dimension of keys $d$.}
    \label{tab:nhead}
    \centering
    \begin{tabular}{c|c|c|ccc}
    \toprule
        \multicolumn{3}{c|}{Settings}  &  \multicolumn{3}{c}{OA} \\ \hline
       $K_3$ & $H$ & $d$ & IP & PU & Houston \\ \hline
        64 & 2 & 32 & 91.40 & 98.78 & 96.12 \\ 
        64 & 4 & 16 & 92.90 & 98.96 & 97.68 \\ 
        64 & 8 & 8 & \textbf{94.14} & \textbf{98.98} & \textbf{98.21} \\ 
        64 & 16 & 4 & 92.08 & 98.86 & 97.82 \\ 
        64 & 32 & 2 & 93.35 & 98.92 & 97.38 \\ 
    \bottomrule
    \end{tabular}
\end{table}

\textbf{Dimension reduction for MPCA.} We adopt the multiview PCA for spectral dimension reduction in the multiview transformer. The output dimension depends on the number of views $g$ and the dimension of each view representation $d$, which is $g\times d$. We first conduct experiments with varying numbers of views $g$ and $d=3$. Results are presented in \cref{tab:abla_nview}. We further explore the effectiveness of dimension $d$ of each view representation with $10$ views. \cref{tab:abla_ncom} shows the OAs on three datasets. We find that the performance in terms of two values varies on three datasets. Since imaging sensors and recorded scenes vary in different HSIs, the optimal settings of different HSIs may be significantly different. Careful parameter design is required for a specific HSI based on its properties. The output of MPCA is expected to have low-dimensional representation while preserving critical information for classification. Therefore, finding the optimal setting of $g$ and $d$ is challenging.

\textbf{Data size for model training.} To further demonstrate the learning capability of the multiview transformer, we present classification results obtained by our proposed method with the different number of training samples on three HSI datasets. The proportion of samples used for model training is $[5\%, 10\%, 15\%, 20\%]$. OAs are reported in \cref{fig:train_size}. As the number of training samples increase, the classification quality (OA) gradually improves. It implies that our proposed model can learn comprehensive and discriminative spatial-spectral features with abundant training samples.

\textbf{Heads and the dimension of queries, keys and values for transformer.} Multi-head attention aims at learning diverse features from inputs. In SPTT, we use the residual connection to integrate features in different layers and accelerate the training process. Therefore, \cref{eqa:khd} should be met when designing the number of 
 heads $H$ and the dimension of keys $d$ for SPTT. OAs of SPTT with different settings are shown in \cref{tab:nhead}. There is a trade-off between $H$ and $d$. Multiple heads $H$ tend to learn rich features, while $d$ decides the expressibility of keys, quires and values. Our method achieves the best performance when $d=H=8$.

\subsection{Dicussion}
Our proposed multiview transformer obtains the best experimental results on three HSI datasets, which demonstrates superiority compared to the existing advanced methods. Our model can learn discriminative feature representations from extremely class-imbalanced training samples. We adopt rigid experimental settings to avoid the spatial overfitting issue, including small patch sizes and limited training samples. Moreover, we conduct experiments on rotated test samples, as shown in \cref{tab:abla_rot}. Results obtained by the multiview transformer on rotated samples achieve similar classification scores on original samples. It implies that our model is less affected by the spatial overfitting problem. However, there are still some limitations in the proposed multiview transformer. Designing the optimal hyperparameters for a specific HSI dataset is challenging. Besides, the labeled categories in existing HSI datasets are limited. Though we evaluate our model on complex real-world datasets, the output quality on an HSI dataset recording a more complex scene is still an open question.

\begin{table}
\caption{Quantitative analysis of a trained multiview transformer on the original and rotated test samples.}
    \label{tab:abla_rot}
    \centering
    \begin{tabular}{c|c|ccc}
    \toprule
        Rotation            & Metrics & IP & PU & Houston \\ \hline
        \multirow{2}*{$0^{\circ}$}    & OA & 94.14 & 98.87 & 98.21 \\  \cline{2-5}
                           ~& AA & 95.34 & 98.26 & 98.40 \\  \hline
        \multirow{2}*{$180^{\circ}$}  & OA & 94.11 & 98.72 & 98.14 \\  \cline{2-5}
                           ~& AA & 94.69 & 97.93 & 98.27 \\  
    \bottomrule
    \end{tabular}
\end{table}

\section{Conclusion}
\label{sec:conclusion}

In this article, we summarized the spatial overfitting issue in the existing HSI classification methods and proposed a multiview transformer with strict experimental settings to avoid this problem. The spatial overfitting issue occurs when using HSI datasets recorded in a simple scene to train a deep learning model. The model is prone to learn scene-specific but not general spatial-spectral correlation, which makes it difficult to properly access the power of a model. We presented possible solutions to avoid this issue and adopted rigorous experimental settings in this paper. Moreover, we introduced the multiview transformer for HSI classification. In the multiview transformer, MPCA is conducted for HSI dimension reduction to yield a low-dimension multiview representation while preserving more spectral details. CNN-based SED extracts a feature cuboid from the multiview representation using a U-shape structure in spectral dimension. The following SPTT efficiently converts the feature cuboid to tokens and utilizes the attention modules to generate discriminative features for classification. Extensive experiments with rigid settings on three HSI datasets and comparisons with existing methods have demonstrated that multiview transformer achieves high-quality outputs and is promising to avoid spatial overfitting problems.

\bibliographystyle{ACM-Reference-Format}
\bibliography{myref}

\end{document}